\documentclass[sigconf]{aamas}

\usepackage{amsmath}

\usepackage{amssymb}
\usepackage{amsthm}
\usepackage{stfloats}

\usepackage{algorithm}
\usepackage{psfrag}
\usepackage{rotating}
\usepackage{gensymb}

\usepackage{subfigure}

\newcommand{\real}{{\mathbb{R}}}

\newcommand{\PP}{{\mathcal{P}}}

\renewcommand{\epsilon}{\varepsilon}

\usepackage{balance} 

\setcopyright{ifaamas}
\acmConference[AAMAS '26]{Proc.\@ of the 25th International Conference
on Autonomous Agents and Multiagent Systems (AAMAS 2026)}{May 25 -- 29, 2026}
{Paphos, Cyprus}{C.~Amato, L.~Dennis, V.~Mascardi, J.~Thangarajah (eds.)}
\copyrightyear{2026}
\acmYear{2026}
\acmDOI{}
\acmPrice{}
\acmISBN{}

\acmSubmissionID{1458}

\title[Analytical Swarm Chemistry]{Analytical Swarm Chemistry: Characterization and Analysis of Emergent Swarm Behaviors}

\author{Ricardo Vega}
\affiliation{
  \institution{George Mason University}
  \city{Fairfax}
  \country{United States of America}}
\email{rvega7@gmu.edu}

\author{Connor Mattson}
\affiliation{
  \institution{Kahlert School of Computing, University of Utah}
  \city{Salt Lake City}
  \country{United States of America}}
\email{c.mattson@utah.edu}

\author{Kevin Zhu}
\affiliation{
  \institution{George Mason University}
  \city{Fairfax}
  \country{United States of America}}
\email{kzhu4@gmu.edu}

\author{Daniel S. Brown}
\affiliation{
  \institution{Kahlert School of Computing, University of Utah}
  \city{Salt Lake City}
  \country{United States of America}}
\email{daniel.s.brown@utah.edu}

\author{Cameron Nowzari}
\affiliation{
  \institution{George Mason University}
  \city{Fairfax}
  \country{United States of America}}
\email{cnowzari@gmu.edu}

\begin{abstract}
Swarm robotics has potential for a wide variety of applications, but real-world deployments remain rare due to the difficulty of predicting emergent behaviors arising from simple local interactions. Traditional engineering approaches design controllers to achieve desired macroscopic outcomes under idealized conditions, while agent-based and artificial life studies explore emergent phenomena in a bottom-up, exploratory manner. In this work, we introduce \textit{Analytical Swarm Chemistry}, a framework that integrates concepts from engineering, agent-based and artificial life research, and chemistry. This framework combines macrostate definitions with phase diagram analysis to systematically explore how swarm parameters influence emergent behavior. Inspired by concepts from chemistry, the framework treats parameters like thermodynamic variables, enabling visualization of regions in parameter space that give rise to specific behaviors. Applying this framework to agents with minimally viable capabilities, we identify sufficient conditions for behaviors such as milling and diffusion and uncover regions of the parameter space that reliably produce these behaviors. Preliminary validation on real robots demonstrates that these regions correspond to observable behaviors in practice. By providing a principled, interpretable approach, this framework lays the groundwork for predictable and reliable emergent behavior in real-world swarm systems.
\end{abstract}


\newcommand{\BibTeX}{\rm B\kern-.05em{\sc i\kern-.025em b}\kern-.08em\TeX}

\begin{document}


\pagestyle{fancy}
\fancyhead{}


\maketitle 


\section{Introduction}

Swarms are purported to be useful in many real-world applications including pollution monitoring~\cite{GZ-GKF-DPG:11}, disaster management systems~\cite{HK-CWF-IT-BS-ER-KP-AW-JW:12}, surveillance~\cite{MS-JC-LP-JT-GL-AT-VV-VK:14}, and search and rescue~\cite{RA-JJ-BA-EM:20}; but after decades of research we see very few, if any, robot swarms being the chosen solution over highly-coordinated multi-robot teams or even single sophisticated robots. This is likely due to our still-limited understanding of swarm control, as the naturally complex interactions among agents make it difficult to predict and manage the emergent behaviors that arise~\cite{JT:05,JT:05ten,OTH:07}.

The Agent-Based Modeling (ABM) and Artificial Life (Alife) communities have a rich history of using simple local rules to uncover complex collective phenomena, ranging from Conway’s Game of Life~\cite{LSS-PES:78} and cellular automata~\cite{SW:83} to more recent studies of flocking and other swarm systems~\cite{HS:25, EG-LH-RN-MM:22, FN-PW-RN:21, CM-DB:23, CM-VR-RV-CN-DSD-DSB:25}. These approaches are often exploratory and bottom-up: researchers specify diverse sets of local interaction rules and observe the emergent macroscopic patterns that arise. 

In contrast, the engineering community typically adopts a top-down perspective, beginning with a desired collective outcome and working backwards to design local rules or controllers that guarantee the emergence of this outcome. For example, control-theoretic and optimization-based methods have been developed to achieve consensus, coverage, formation control, and cooperative search~\cite{CT-CL-CN:20, FB-JC-SM:09, CN-JC:11-auto, JC-SM-TK-FB:02-tra, ROS-JAF-RMM:07}. There, the focus is on analysis, synthesis, and guarantees: given a well-defined system-level goal, how can one find local interaction strategies that provably achieve the desired macroscopic effect? By considering different sensing and perception models, many different methods of such controllers can be designed. This perspective emphasizes tractability, scalability, and the practical realities of deployment in robotic systems, where reliability and performance are paramount.

Through this traditional engineering approach, an ``optimal” deployment strategy is identified, one that specifies a particular controller for a given number of robots with defined capabilities to optimize key performance metrics. However, in real-world deployments, unforeseen issues frequently arise that can invalidate this ``optimal” configuration; for example, a robot may fail to boot properly or have a faulty wheel. In such cases, what should the next step be? Should the remaining functioning robots be deployed with the same controller in the hope that the system still performs adequately? How can we determine what will actually work without re-running simulations on the spot to search for a new “optimal” configuration under the current conditions?

To address this challenge, we adopt a formal framework for studying and designing robot swarms that integrates analytical methods from engineering with exploratory approaches from the agent-based modeling and simulation communities. Furthermore, inspired by concepts from chemistry, we examine how different system compositions and interaction patterns naturally give rise to distinct macrostates through emergent self-organization.

Connections between swarms and chemistry are not entirely new: the idea of ``swarm chemistry" was first coined by Sayama in 2009 in an Alife article~\cite{HS:09} where it was found that mixing agents with different control rules led to fascinating new behaviors. However, these swarm chemistry ideas primarily only consider combining two or more different ``elements" or ``species" together~\cite{HS:09, HS:11, HS:12, HS:25}. We believe the connection to chemistry goes much deeper and exploring these ideas for even homogeneous agents is a missed opportunity. We can borrow the use of phase diagrams from chemistry as a key tool to better understand and visualize the phases of the swarm. In chemistry, phase diagrams document how the state of water (solid, liquid, gas) transitions as a function of pressure and temperature~\cite{BP-MH-MJP:13}. 

In our work, we investigate existing controllers in the literature and examine how a series of parameters impacts the emergent properties of the swarm. A major difference is that we are dealing with a much higher dimensional parameter space as we are dealing with sensing and acting agents rather than inanimate water molecules. However, the question remains fundamentally the same: can we identify and visualize the subsets of a swarm parameter space to allow us to better understand when a desired behavior self-organizes without running new simulations every time? 


This work contributes a new framework, \textit{Analytical Swarm Chemistry}, which integrates macrostate definitions and phase diagram analysis to systematically explore the relationship between swarm parameters and emergent behavior. By applying the framework to minimal agent models, we show how it can reveal sufficient conditions for desired behaviors --- laying groundwork for future extensions to more complex swarm systems, some of which we preliminarily validate using real robots.


\section{Problem Formulation}\label{se:problem_formulation}

Consider a very simple swarm of~$N$ self-propelled agents moving in a 2D environment~$\mathcal{D}~\subset~\real^2$. The 2D position and orientation of each robot at time $t$ is given by $\mathbf{q}_i(t) = [x_i(t),y_i(t)]^T \in \mathcal{D}$ and~$\theta_i(t) \in [-\pi,\pi)$, respectively, such that the full observable state of agent $i$ is $\mathbf{p}_i(t) =  [\mathbf{q}_i(t), 
\theta_i(t)]^T \in \PP=\mathcal{D} \times [-\pi,\pi)$ with 
\begin{eqnarray}\label{eq:simple_kinematics}
     \left[ \begin{array}{c} \dot{x}_{i}(t) \\ \dot{y}_{i}(t)\\ \dot{\theta}_{i}(t)\end{array} \right]  = f_i(\mathbf{p}_i(t),\mathbf{u}_i(t)) =\left[ \begin{array}{c} u_{i,1}(t)\cos \theta_{i}(t) \\ u_{i,1}(t) \sin \theta_{i}(t) \\ u_{i,2}(t)  \end{array} \right].
\end{eqnarray}

These agents have a single, forward-facing binary sensor that is triggered when at least one other agent is within the sensor's field of view, which is the conical area in front of the robot with range~$\gamma>0$ and opening angle~$\phi>0$ as shown in  Figure~\ref{fig:behaviors_figure} and denoted as $\operatorname{FOV}_i$ for agent $i$:
\begin{eqnarray}\label{eq:mill_output}
    h_i = \begin{cases} 
                 1 & \text{ if } \exists j \neq i, s.t. ~\textbf{q}_j \in \operatorname{FOV}_i , \\
                 0 & \text{otherwise.}
             \end{cases}
\end{eqnarray}

The control input of the agents $u_i(t)$ consists of a speed and turning rate $u_i  = \left[ v , \omega \right]^T$ and is a function of their sensor output $h_i(t)$.

There has been multiple studies investigating what behaviors can emerge given these limited, binary agents~\cite{MG-JC-TJD-RG:14, MG-JC-WL-TJD-RG:14, AO-MG-AK-MDH-RG:19, FB-MG-RN:21, DS-CP-GB:18, DB-RT-OH-SL:18, CM-DB:23, CM-VR-RV-CN-DSD-DSB:25}. These works primarily focus on the interaction rules between the agents and mainly explore the controller $u_i(t)$ and how different control laws can result in different macro behaviors. However, our goal here is \textbf{not} to design specific control inputs to induce some desired behaviors; rather we attempt to identify and characterize the conditions that lead to different macro-behaviors and the macroscopic properties applicable therein. 

We specifically look at the conditions of the parameter space that elicit or inhibit the behavior from being produced. We denote $\mathcal{R}$ as the parameter space that includes the parameters that describe the system such as the system parameters (e.g., number of agents), the environment parameters (e.g., size, obstacle density), and the agent parameters (e.g., sensing and actuating limits). Clearly, two systems operating under the same control law $u_i$ may result in dramatically different end behaviors (e.g. 100 robots with long, narrow sensing region vs 4 robots with short-distance 360\degree ~sensors). 

Rather than focusing on finding a single optimal point (as is done traditionally), our objective is to identify regions within $\mathcal{R}$ where a given behavior $B$ reliably emerges, even if not optimally. This leads to the central question of this work:

\textit{Given a control law $u_i$ known to sometimes produce a behavior $B$, can we determine the regions in $\mathcal{R}$ where $B$ occurs reliably?}

\section{Analytical Swarm Chemistry}\label{se:swarm_chem2}
In this section, we present the proposed framework, which takes the first steps toward addressing the question posed in Section~\ref{se:problem_formulation} by identifying subsets of the parameter space where the desired behaviors consistently emerge. The framework does not seek to exhaustively map all regions of $\mathcal{R}$ in which a behavior may occur. Instead, it aims to uncover conditions that are sufficient for producing the behavior --- while acknowledging that other, yet undiscovered, regions of $\mathcal{R}$ may also support it.

\subsection{Defining Macrostates}

The first half of our framework utilizes information markers to define a macrostate that we can use to classify the swarm behavior. To define a macrostate of a swarm, we look towards statistical mechanics and thermodynamics where macrostate is often defined as being a set of microstates that share specific macroscopic properties \cite{CRS-CM:25}. Here we use the term microstates to refer to the collective observable states of all the agents.  More formally, we define this microstate as~$P = (\mathbf{p}_1, \dots, \mathbf{p_N}) \in \mathcal{P}^{N} $. 

Regardless of the controllers used, we can start with objectively measurable properties of the trajectories of all agents~$P(t)$. Just as thermodynamics uses the well established notions of temperature, pressure, energy, density, and more to group microstates of a molecular system into macrostates, we can look at different measurable properties to identify exactly how the behaviors of the swarm may differ.  Although, \cite{HH-KJ-JL:21} identified and defined the macroscopic properties they called `swarm temperature', `swarm pressure', and `swarm density'; their system was based on attractive/repulsive agent behaviors similar to real molecular dynamics. As a result, it cannot be readily generalized to other types of swarm systems such as ours. In contrast, our framework is designed to explore and identify behaviors across a broader range of swarm models.

Following previous works that have established frameworks that allow swarm systems to be more formally analyzed~\cite{RV-CN:25, AJH-AH-DJR-HAA:23, AJH-ASMH-DJR-HAA:24}, we let~$F: \mathcal{P}^N \rightarrow \mathcal{Y}$ be a map that processes the observable data into the set of all information~$\mathcal{Y}$ with elements~$Y_\ell = F_\ell(P)$, essentially representing the measurable macroscopic properties of the microstate. This information set encompasses all the ways we can measure what occurs in the system at a high-level; as there are infinite different metrics that can be measured, this information set $\mathcal{Y}$ is infinitely large. An information marker is a subset of information~$M = G(Y) \in \real^{m}$ where, in general, these are reduced-order measurable outputs~$G: \mathcal{Y} \rightarrow \real^{m}$ of the entire state~$P(t)$ to summarize the information contained within the entire system to a few metrics of interest based on the desired behavior.

We then define a macrostate as the group behavior~$B_j$ which occurs if its associated information markers~$M^j \in \real^{m^j}$ belong to the set~$\eta^j \subset \real^{m^j}$, the structure composing the behavior. That is, if~$M^j\in\eta^j$, then the system is in macrostate~$B_j$. 
\begin{eqnarray}\label{eq:farp_output}
    B_j = \begin{cases}
                 1 & \text{if }M^{j} \in \eta^{j} , \\
                 0 & \text{otherwise.}
             \end{cases}
\end{eqnarray}

There may be multiple ways to define when a behavior occurs  depending on the chosen information markers. However, changing these markers necessitates a different structure set $\eta$. As long as the selected markers fall within their corresponding structure set, the behavior can be considered validly produced. Similarly, by altering the definition of the structure set, it is possible to define different macrostates using the same information markers as those associated with another behavior.

By defining macrostates in this manner, we can objectively evaluate whether a behavior is produced within the system. The choice of information and behavior markers is selected by the user: markers are selected such that, when the values fall within the user-specified structure set, it is clearly evident that the behavior is occurring. These selections are informed by the user's understanding of the system and intuition about what constitutes the behavior. Once appropriate markers and thresholds are established, this approach allows us to systematically explore the conditions, specifically, the parameter sets, that give rise to the specified macrostates.

\subsection{Visualizing Macrostates with Phase Diagrams}\label{sse:phase_approach}
The aim of this section is to provide tools for characterizing emergent behaviors in swarms of agents. Just as chemists use phase diagrams to predict the state of matter of a substance under specific conditions, we can construct analogous diagrams to visualize how different parameter combinations lead to distinct macrostates in a swarm system. However, unlike in chemistry, the macroscopic properties of a swarm cannot be directly used as diagram axes due to the added agency and nonlinear interactions introduced by even simple sensor-to-actuator control. Instead, we construct phase diagrams using system parameters that influence these macroscopic properties. While similar diagrams have been explored in previous simulation-based studies to identify behavioral regimes across controller or rule parameters~\cite{AC-CKH:18, MRD-YC-ALB-LSC:06, NVB-HA-IYT-SAM:20, ZC-ZC-VT-DC-HZ:16}, our framework extends this concept by systematically exploring macrostates over the larger underlying parameter space.

From Section~\ref{se:problem_formulation}, several parameters can be identified that describe the system. In this case, we consider the parameters as independent variables that can be adjusted, including the number of agents~$N$, agent speed~$v$, turning rate~$\omega$, vision distance~$\gamma$, and field-of-view opening angle~$\phi$. Since the environment is open and contains no obstacles, environmental parameters can be neglected.

\begin{figure}[b]
\centering
  \subfigure[]{\includegraphics[width=.49\linewidth]{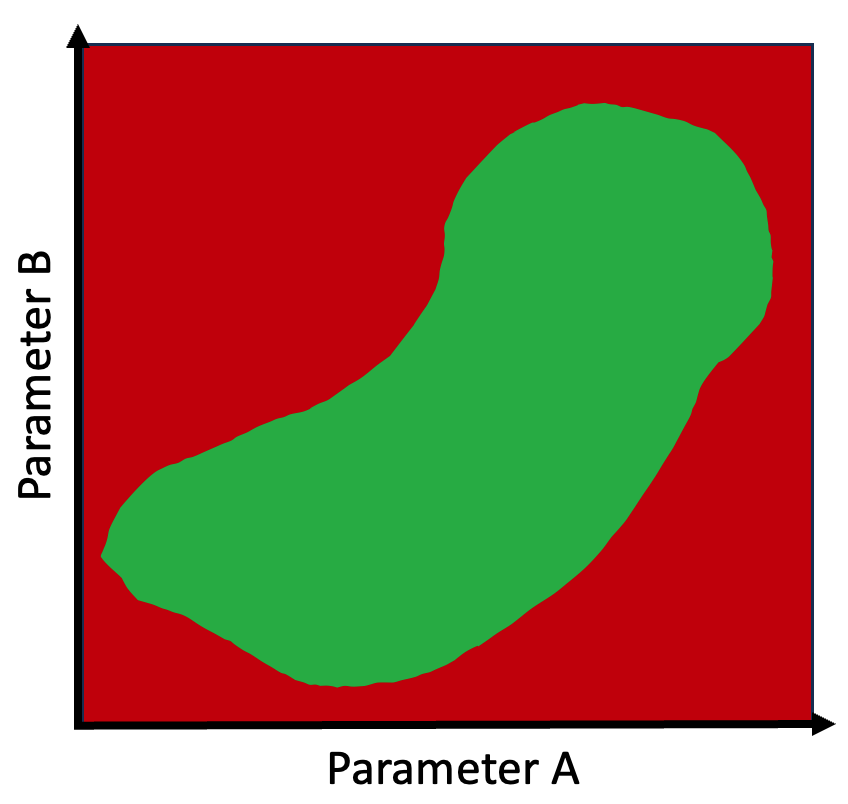}}
  \subfigure[]{\includegraphics[width=.49\linewidth]{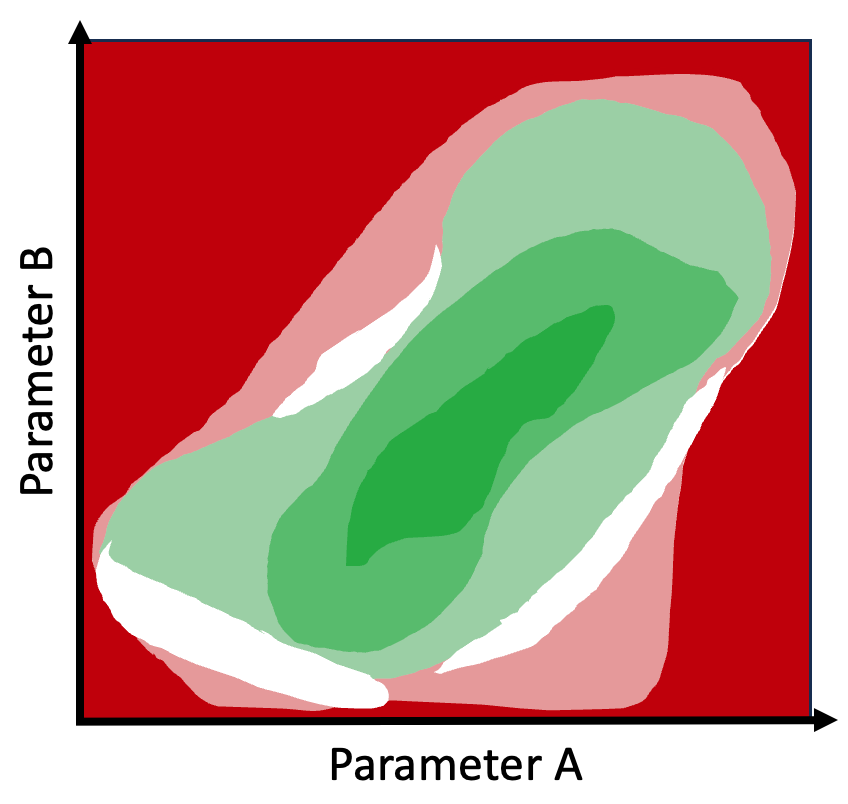}}
  \caption{(a) Abstract phase diagram showing where $B_j = 1$, depicted as the green area, in respect to Parameter A and B. (b) Abstract phase diagram that uses intensity of color to represent how frequently behavior $B_j$ occurred out of the total runs simulated.}
\label{fig:abstract_phase}
  \Description{Phase diagram of an abstract example}
\end{figure}

By simulating the system across different parameter combinations, we can identify where a behavior of interest occurs. This concept is illustrated in the abstract example shown in Figure~\ref{fig:abstract_phase}(a). In this case, the phase diagram represents two possible outcomes: the macrostate $B_j$ either occurs ($B_j = 1$) or it does not ($B_j = 0$). This approach allows us to concentrate our analysis on regions of the parameter space where the desired behavior reliably emerges, rather than on other possible macrostates.

When constructing these phase diagrams, it is essential to perform multiple simulation runs for each parameter combination to ensure that the behavior occurs consistently, rather than as a one-off event. Variability may arise if the initial conditions are slightly unfavorable in one run but conducive in others. To visualize this reliability, we adjust the color intensity in the phase diagram according to how frequently the system produces the behavior across repeated trials,as shown in Figure~\ref{fig:abstract_phase}(b). 

For example, if each parameter combination between parameters A and B are simulated ten times, a solid green region indicates that the example behavior occurred ($B_j = 1$) in every trial. A lighter shade of green denotes where the behavior emerged in a slight majority of runs (e.g. six or seven out of ten). Conversely, if the behavior failed to occur in the majority of trials ($B_j = 0$), the region is shown in a light red shade, while consistent failure across all runs is represented by a solid dark red. If there is no clear majority outcome (i.e., an equal number of successes and failures), the region is colored white to indicate the absence of a dominant behavior.

Sweeping through all parameters independently produces a multi-dimensional phase space, making direct visualization challenging since any phase diagram represents only a two-dimensional slice of this space. This highlights the vastness of the full search space, even for a relatively simple swarm, and the inherent difficulty of efficiently predicting when specific behaviors will emerge. Nevertheless, phase diagrams serve as a practical and insightful tool for revealing dominant pairwise relationships between parameters and the transition boundaries between behaviors. Currently, selecting cross sections relies on initial simulator runs, domain knowledge, and observed sensitivities, enabling the identification of meaningful parameter relationships and guiding further exploration. Future extensions could incorporate automated techniques, such as principal component analysis or variance-based sensitivity analysis, to systematically identify the most influential parameter combinations and enhance scalability.

\section{Case Studies}\label{se:case_studies}

\begin{figure}[ht]
    \centering
    \includegraphics[width=0.99\linewidth]{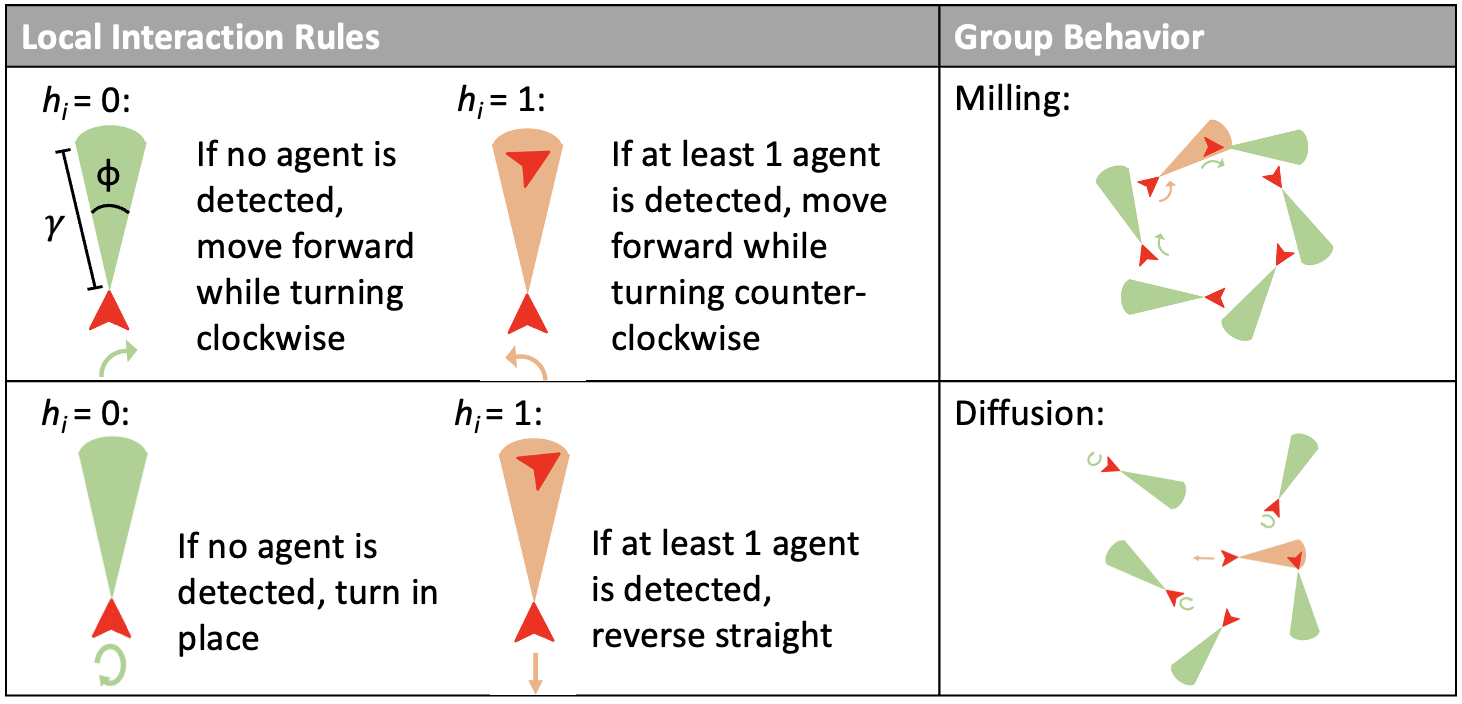}
    \caption{Examples of local interaction rules and their resulting emergent behaviors. The top row shows the controller~\eqref{eq:mill_control} and the corresponding milling behavior it can produce, while the bottom row shows the controller~\eqref{eq:diff_controller} and its resulting diffusion behavior.}
    \Description{Different local interaction rules leading to different emergent group behaviors.}\label{fig:behaviors_figure}
\end{figure}

In this section, we further apply our swarm analytics framework to see how we can help answer the problem from Section~\ref{se:problem_formulation} for two particular canonical behaviors of interest: $B_1 =$ \textit{milling} and~$B_2=$ \textit{diffusion}. For both cases we consider very simple binary sensor-to-action controllers that have previously been discovered (shown in Figure~\ref{fig:behaviors_figure}).

Milling, sometimes referred to as cyclic pursuit, is a commonly studied behavior in which agents organize into a well-formed, rotating circle~\cite{QW-YW-HZ:16, CW-GX:17, XY-LL:17, RZ-ZL-MF-DS:15, SY-SP-YK:13, CT-CL-CN:20}. This behavior is often examined because it presents a moderately complex coordination challenge while remaining easily recognizable in both simulation and physical systems.

The local controller considered in this work, which can produce milling under appropriate conditions, is given by:
\begin{eqnarray}\label{eq:mill_control}
   u_i(t) = \begin{cases}
                 [v, \omega]^T & \text{if } h_i(t) = 1 ,\\
                 [v, -\omega]^T & \text{otherwise,}
             \end{cases}
\end{eqnarray}
where~$v > 0$ and~$\omega > 0$ are the selected forward speed and turning rate of all the agents, respectively. This simple binary controller was chosen as it was found to be capable of producing the emergent milling behavior given these limited capabilities in past works~\cite{MG-JC-TJD-RG:14, DB-RT-OH-SL:18, DS-CP-GB:18, FB-MG-RN:21}. However, this milling macrostate arises only under specific, nontrivial conditions that have not yet been clearly characterized. As such, a detailed analysis is warranted to better understand the conditions that give rise to milling.

Diffusion, closely related to spatial coverage, is another commonly studied behavior in swarm systems~\cite{QW-HZ:21, AO-MG-AK-MDH-RG:19, XL-YT:17, HO-YJ:14}. In this paper, diffusion refers to the behavior where agents start in close proximity and spread out \textit{evenly}, not merely moving away from each other, but arranging themselves so that the distances to their nearest neighbors are approximately equal, resulting in low variance. This behavior is particularly useful in applications that require rapid and uniform coverage, such as search and rescue operations. To generate this behavior given our system, we use the binary local controller described in~\cite{AO-MG-AK-MDH-RG:19}:
\begin{equation} \label{eq:diff_controller}
\begin{split}
u_{i}(t) &= \begin{cases} (-v,0) \quad &\text{if } h_i(t) = 1 , \\ (0,\omega) \quad &\text{otherwise.} \end{cases} 
\end{split}
\end{equation}

The remainder of this section shows how to objectively formalize the two macroscopic behaviors above using the analytical swarm chemistry framework.

\subsection{Defining Macrostates}
In order to define when the behavior occurs, we first must find what measurable properties/information should be measured. 




\begin{table*}[t]
  \caption{Information markers measurable by an external observer. This is not an exhaustive list, any markers could be chosen depending on the behavior being defined. The markers listed here are those used to define the milling and diffusion macrostates.}
  \label{tab:properties}
  \begin{tabular}{p{3.5cm} p{1.2cm} p{7.5cm}}\toprule
    \textit{Name} & \textit{Variable} & \textit{Equation} \\ \midrule
    Average  Speed & $Y_1 =\overline{v}$ & $ \frac{1}{N} \sum_{i=1}^{N} ||\dot{\mathbf{p}}_i||_2 $ \\ 
    Circliness & $Y_2 =\overline{c}$ & $\frac{\max_{i \in N} ||\mathbf{p}_i-\mu|| - \min_{i \in N}||\mathbf{p}_i-\mu||}{\min_{i \in N}||\mathbf{p}_i-\mu||}$  \\ 
    Nearest Neighbor Variance & $Y_3 =\overline{\delta}$ & $\frac{1}{ N} \sum_{i=1}^{ N}( \min_{j \neq i}||\mathbf{p}_i - \mathbf{p}_j|| -\frac{1}{ N} \sum_{i=1}^{ N} \min_{j \neq i}||\mathbf{p}_i - \mathbf{p}_j||)^2$  \\ 
    \bottomrule
  \end{tabular}
  \end{table*}

\subsubsection{Milling:}
\begin{figure}[h]
\centering{\includegraphics[width=.7\linewidth]{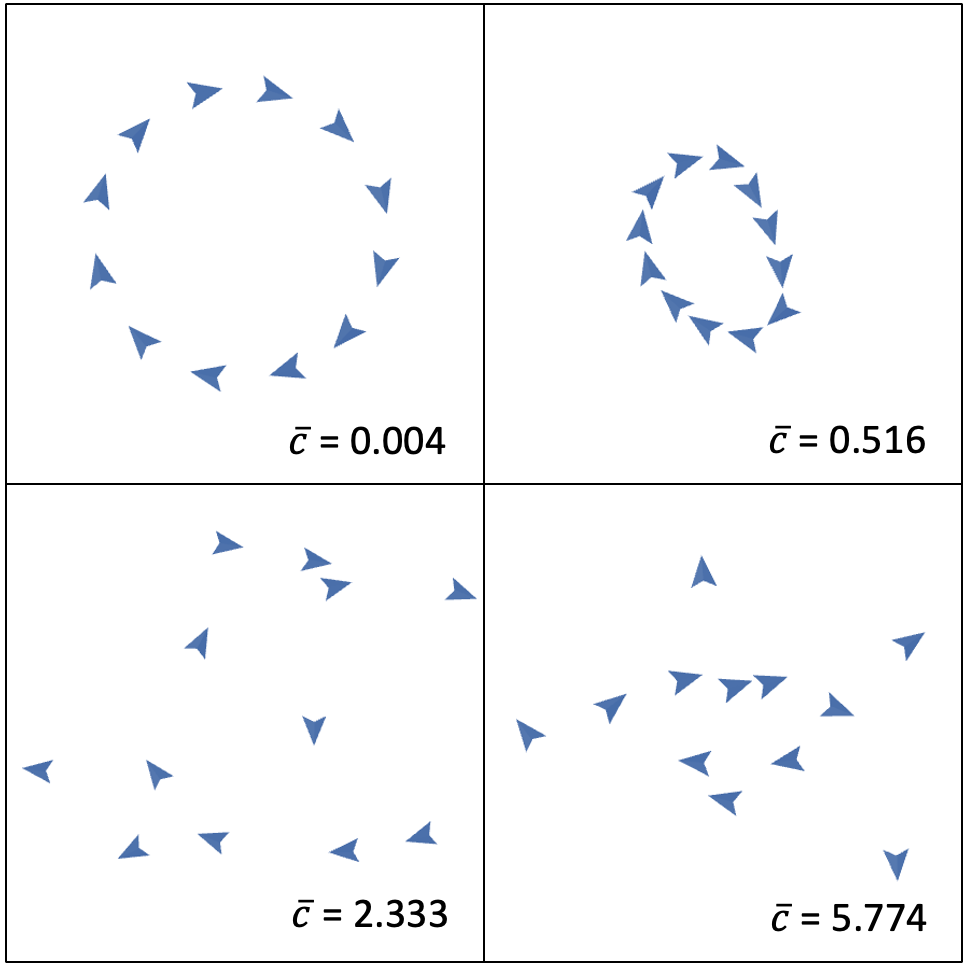}}
  \caption{Examples of different circliness values ($Y_2 = \overline{c}$) used to define the milling behavior. If we assume that the average speed is equal to the set speed of the system $v$ (i.e. the agents are moving constantly), then only the top-left snapshot would be considered milling, as $\overline{c} = 0.004$ therefore $M^1 \in \eta^1$, satisfying the criteria for this behavior.}
  \Description{Examples of different circliness values ($Y_2 = \overline{c}$) used to define the milling behavior.}\label{fig:circliness_metrics_plots}
\end{figure}

For milling, we measure the system’s ``circliness,” similar to the metric in~\cite{CT-CL-CN:20}, which compares the distances of the closest and farthest agents from the center of mass, $\mu(t) = \frac{1}{N}\sum_{i=1}^{N} \mathbf{p}_i(t)$. A value of $\overline{c} = 0$ represents a perfect circle. Figure~\ref{fig:circliness_metrics_plots} illustrates snapshots corresponding to different $\overline{c}$ values. The circliness value $\overline{c}$ should remain near zero; larger values indicate deviations from circularity, which violates our milling definition. In addition, we monitor the system’s average speed, which should match the set forward speed $v$ to ensure that agents remain in motion, as required by our definition of milling. 

We then define the behavior marker of milling, or the macroscopic properties of interest, as $
M^{1} =  [ Y_1 ,Y_2 ]^ T$.
We then defined the structure set for milling as
\begin{eqnarray}
\eta^{1} = {\{(\bar{v} , \bar{c}) \in \real^2 | \bar{v} = v, \bar{c} < 0.01 \}}. 
\end{eqnarray}

These values were chosen empirically by the authors such that if $M^{1}\in \eta^{1}$, any observer would recognize the behavior as milling. 
\begin{eqnarray}\label{eq:mill_behavior_output}
    B_{1} = \begin{cases}
                 1 & \text{if }M^{1} \in \eta^{1} , \\
                 0 & \text{otherwise.}
             \end{cases}
\end{eqnarray}

\begin{figure*}[h!]
\centering
  \subfigure[]{\includegraphics[width=.32\linewidth]{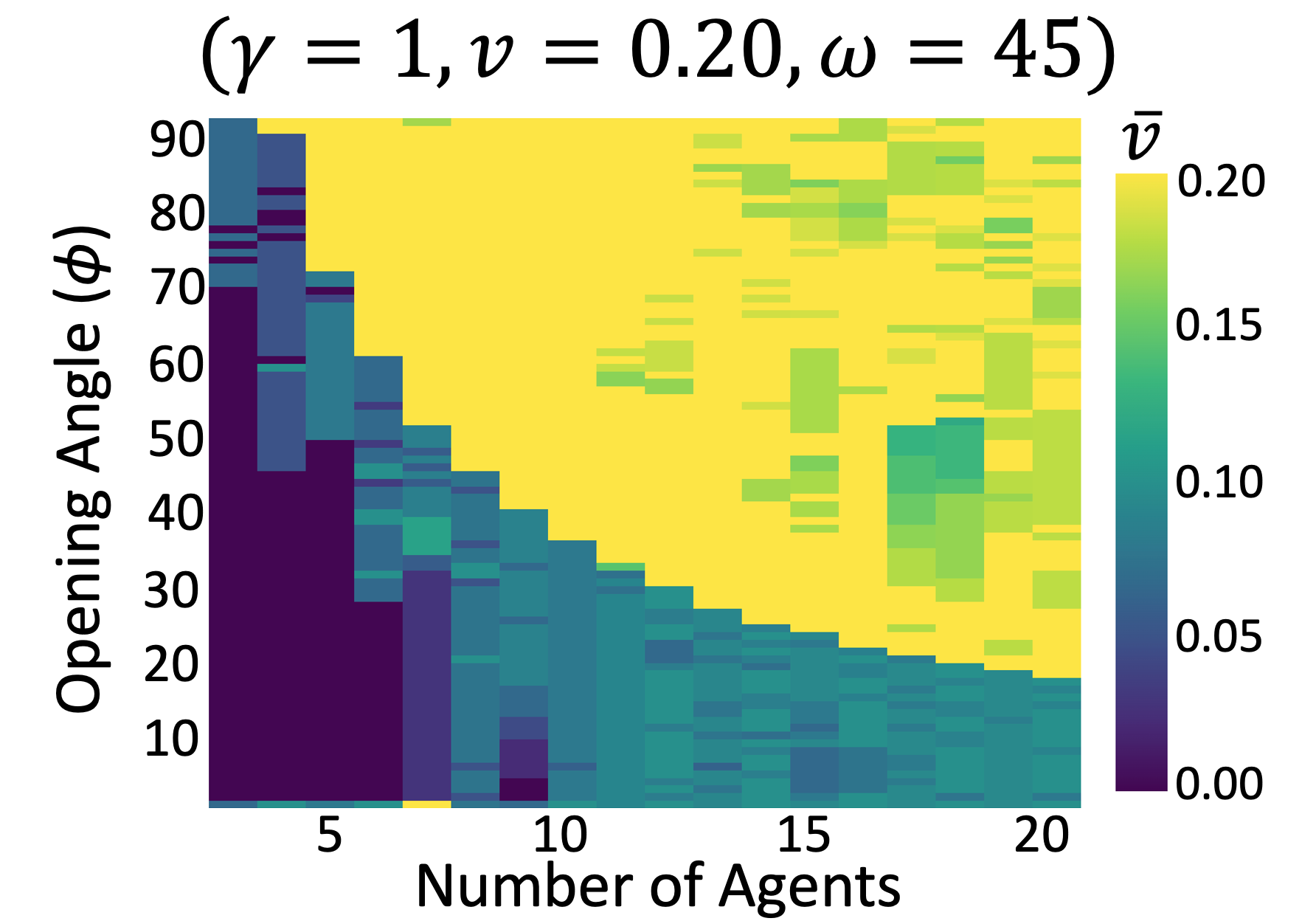}}
  \subfigure[]{\includegraphics[width=.32\linewidth]{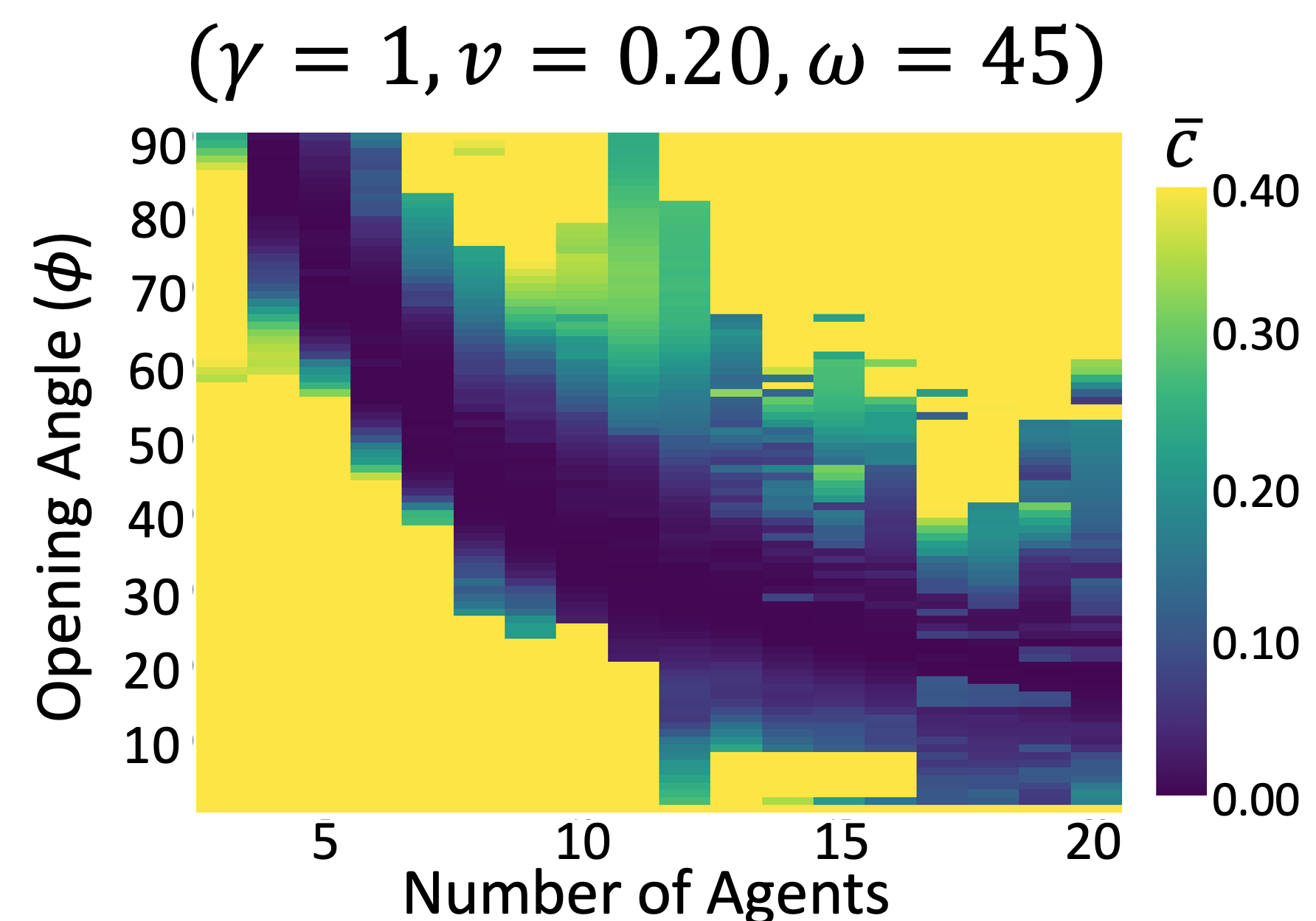}} 
  \subfigure[]{\includegraphics[width=.32\linewidth]{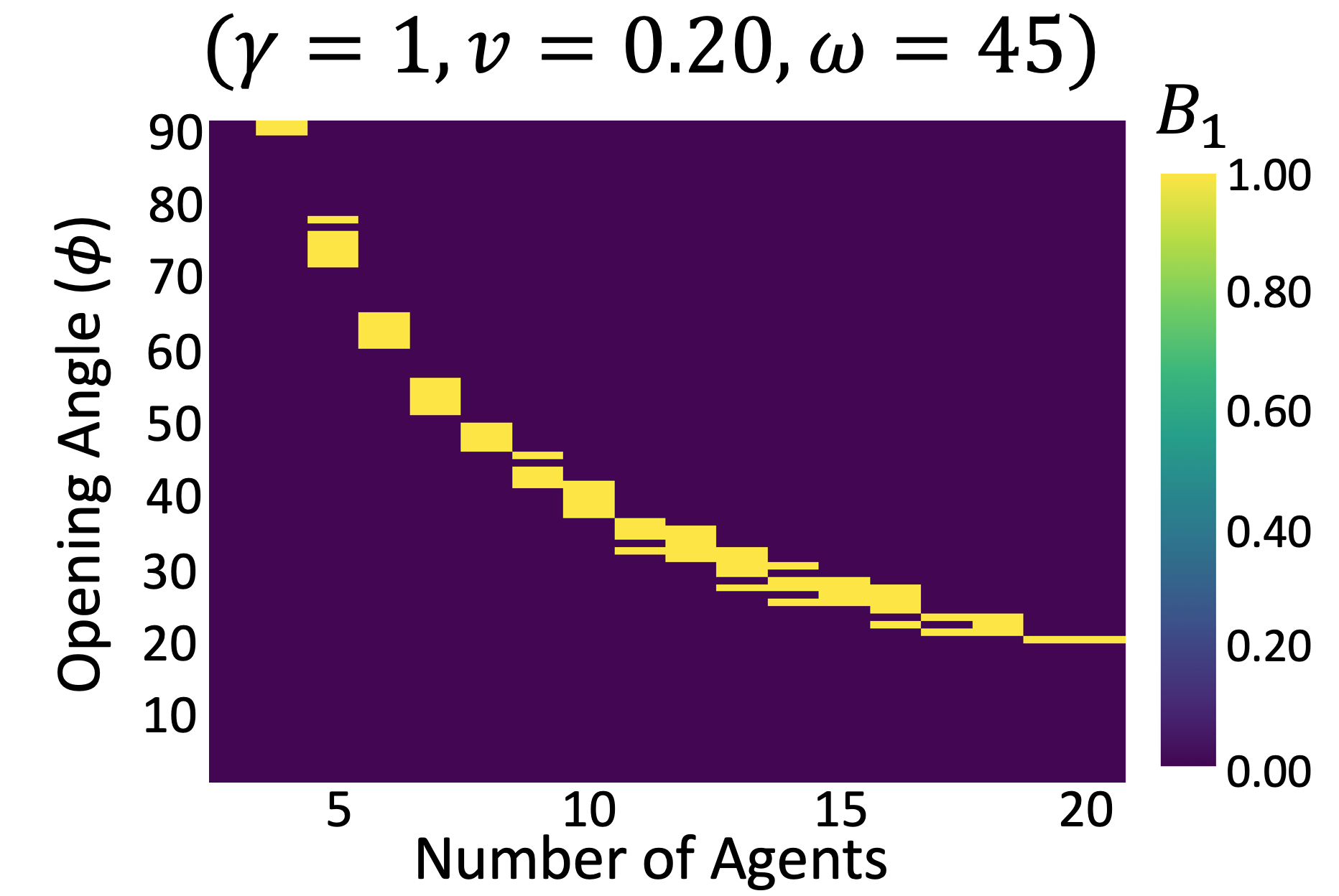}}
  \caption{Phase diagrams of number of agents $N$ vs FOV angle $\phi$ showing: (a) average speed $Y_1 = \bar{v}$, (b) circliness values $Y_2 = \bar{c}$, (c) $B_{1}$ value. (a) and (b) display some important region in circliness and average speed, respectively, but it is made even more clear when both are considered together and compared to $\eta^{mill}$ to identify when milling is occurring. }
  \label{fig:milling-behavior-marker}
\end{figure*}

Figure~\ref{fig:milling-behavior-marker} shows heat maps of the average speed in (a) and circliness values in (b) across various parameter combinations. Some inverse relationships between $N$ and $\phi$ are apparent when examining each graph individually. However, these relationships become much clearer when both information markers are considered together, observing only the regions where $B_1 = 1$, that is, when milling occurs according to our macrostate definition. This illustrates the utility of using multiple information markers simultaneously to more accurately define and detect behaviors.

\subsubsection{Diffusion:}
\begin{figure}[h]
\centering{\includegraphics[width=.7\linewidth]{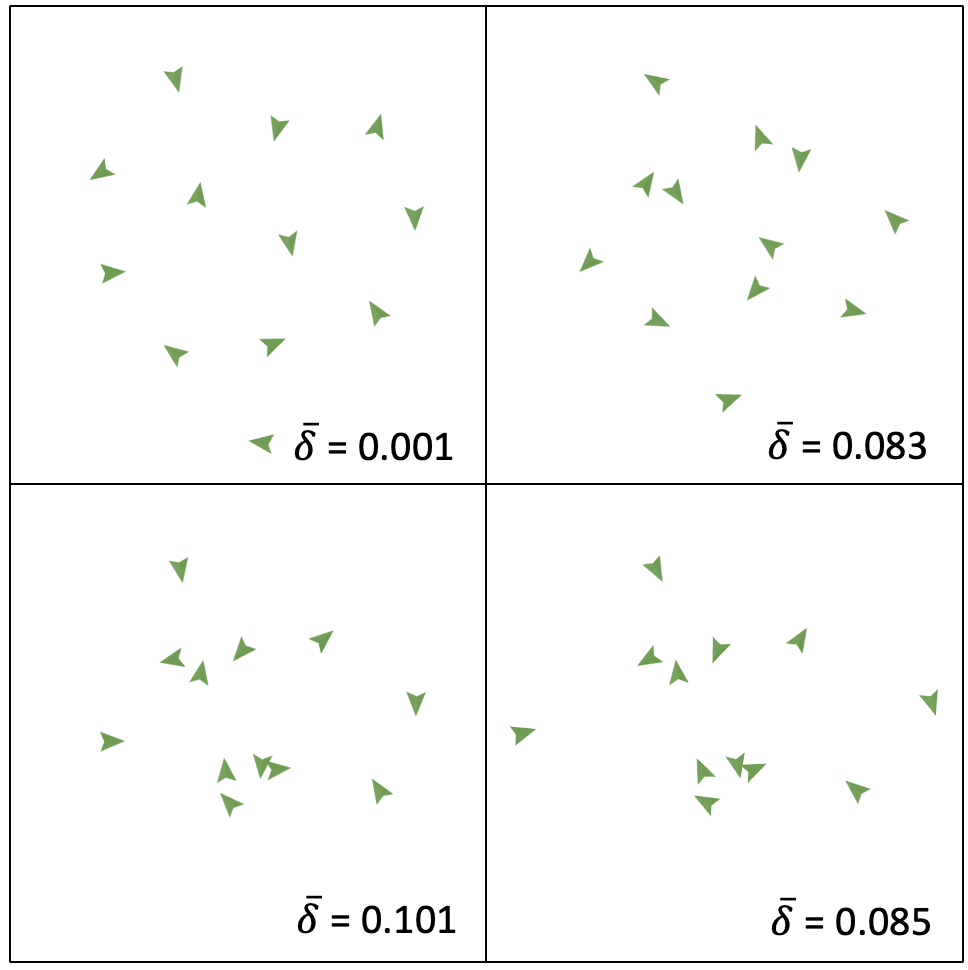}} 
  \caption{Examples of different nearest-neighbor variance values ($Y_3 = \overline{\delta}$) used to define the diffusion behavior. Only the top-left snapshot is considered diffusing, as $\overline{\delta} = 0.001 \in \eta^2$, satisfying the criteria for this information marker.}
  \Description{Examples of different nearest-neighbor variance values ($Y_3 = \overline{\delta}$) used to define the diffusion behavior.}\label{fig:diffusion_metrics_plots}
\end{figure}

Similar to how circliness metric helps identify when the milling behavior occurs, we can use the variance of the minimum distance between neighbors as a separation uniformity metric $\overline{\delta}$. With this metric, we can better identify the diffusion behavior from any dispersal behavior that would be considered the same using something like the scatter metric (i.e. average distance from center of mass). Snapshots of what different values of $\overline{\delta}$ looks like are shown in Figure~\ref{fig:diffusion_metrics_plots}.

As we did for milling, we can now define the behavior marker for diffusion as $M^{2} = Y_3$, and the structure set as
\begin{eqnarray}
\eta^{2} = {\{\bar{\delta}  \in \real | \bar{\delta} < 0.005 \}}
\end{eqnarray}
such that 
\begin{eqnarray}\label{eq:diffuse_behavior_output}
    B_{2} = \begin{cases}
                 1 & \text{if }M^{2} \in \eta^{2} , \\
                 0 & \text{otherwise.}
             \end{cases}
\end{eqnarray}



\subsection{Visualizing Macrostates with Phase Diagrams}\label{se:phase_diagrams}
To create the phase diagrams for each behavior, simulations of the system were run in Netlogo \cite{UW:99}, where the time discretizations was reduced enough such that reducing it further had no change in behavior in order to emulate the continuous-time model~\eqref{eq:mill_control} and ~\eqref{eq:diff_controller} used. It should also be noted that all the simulations were initially setup such that r-disk graph $\mathcal{G}_\text{disk}$, where the edges exist between agents if they are within $\gamma$ of each other, was strongly connected.



\subsubsection{Milling}

After conducting multiple parameter sweeps across $\mathcal{R}$, we observed that systems running the same binary controller can exhibit a range of distinct behaviors, depending on the parameter values. This finding underscores the importance of systematic analysis for real-world swarm applications, as it reveals how sensitive collective behavior can be to underlying system parameters. In our experiments, the binary controller~\eqref{eq:mill_control} produced six qualitatively different outcomes: a well-formed and stable milling circle, an imperfect ellipsoidal formation, a clustered aggregation resulting from collisions, and configurations where the system fragmented into multiple sub-groups. These observations motivate the use of our proposed framework, which enables clear identification of regions in the parameter space where the same control law~\eqref{eq:mill_control} leads to markedly different emergent behaviors. Here, we focus specifically on analyzing the conditions under which stable milling behavior occurs.

\begin{figure*}[h]
    \centering
    \subfigure[]{\includegraphics[width=.33\linewidth]{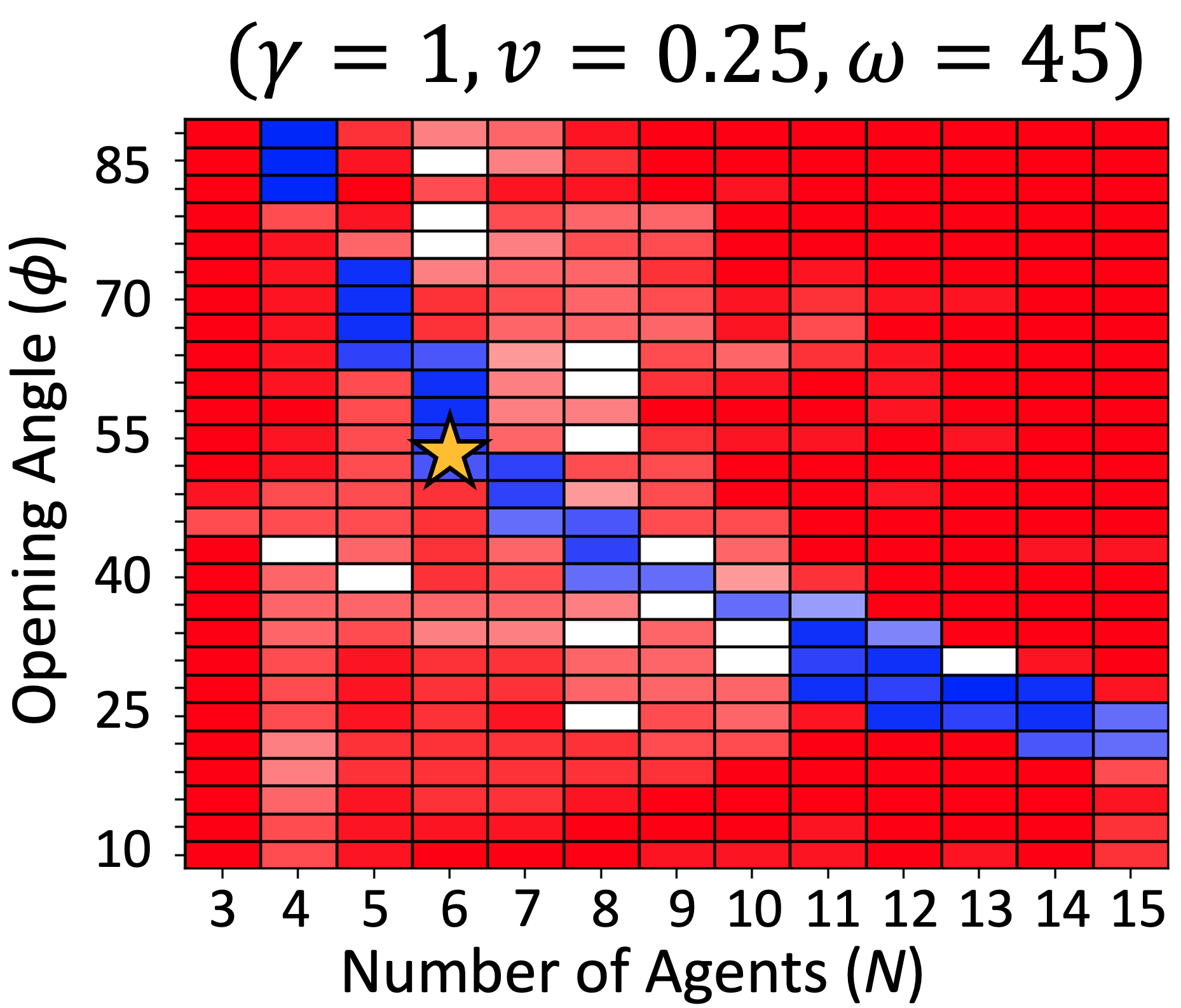}}
  \subfigure[]{\includegraphics[width=.33\linewidth]{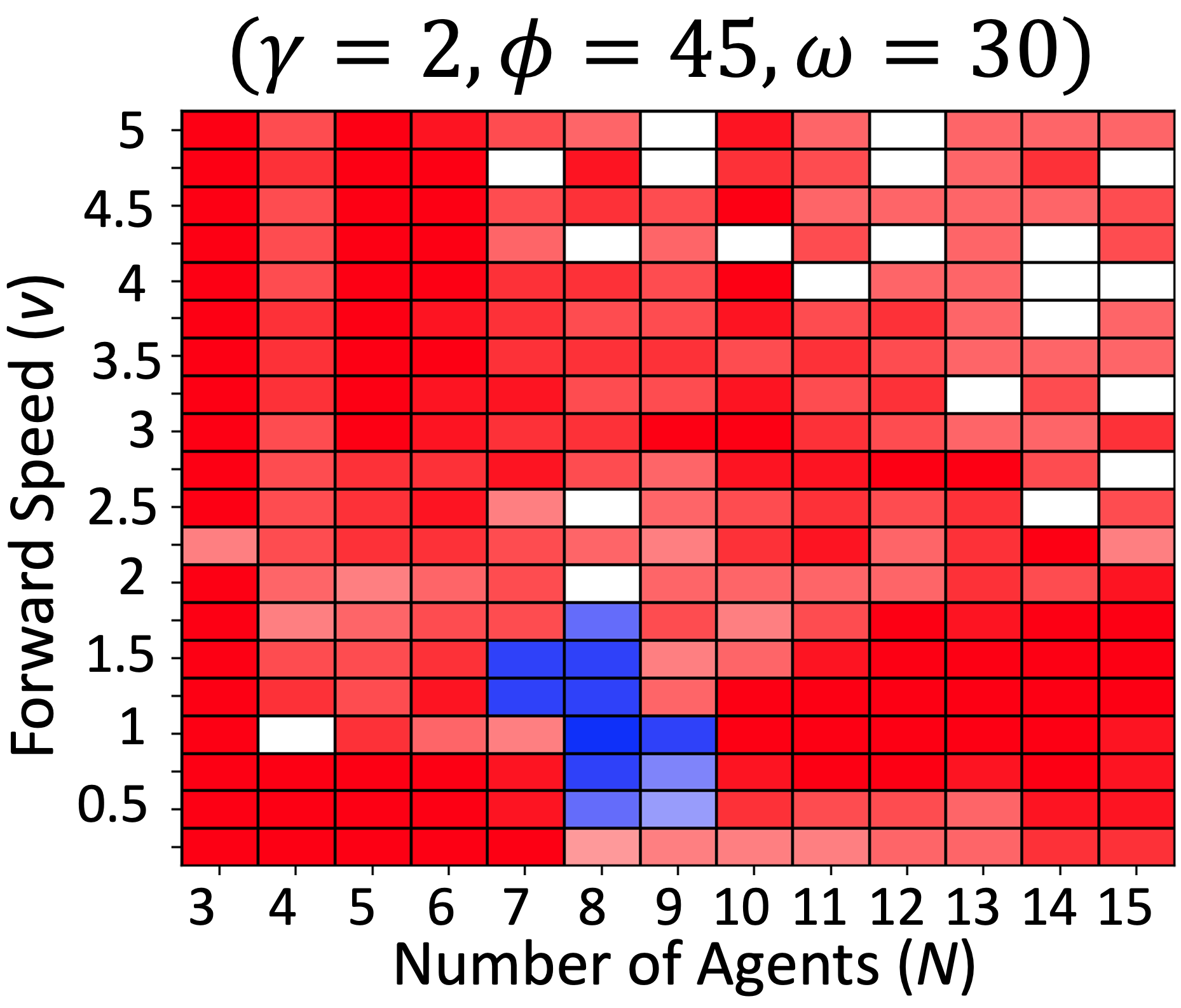}}
  \subfigure[]{\includegraphics[width=.33\linewidth]{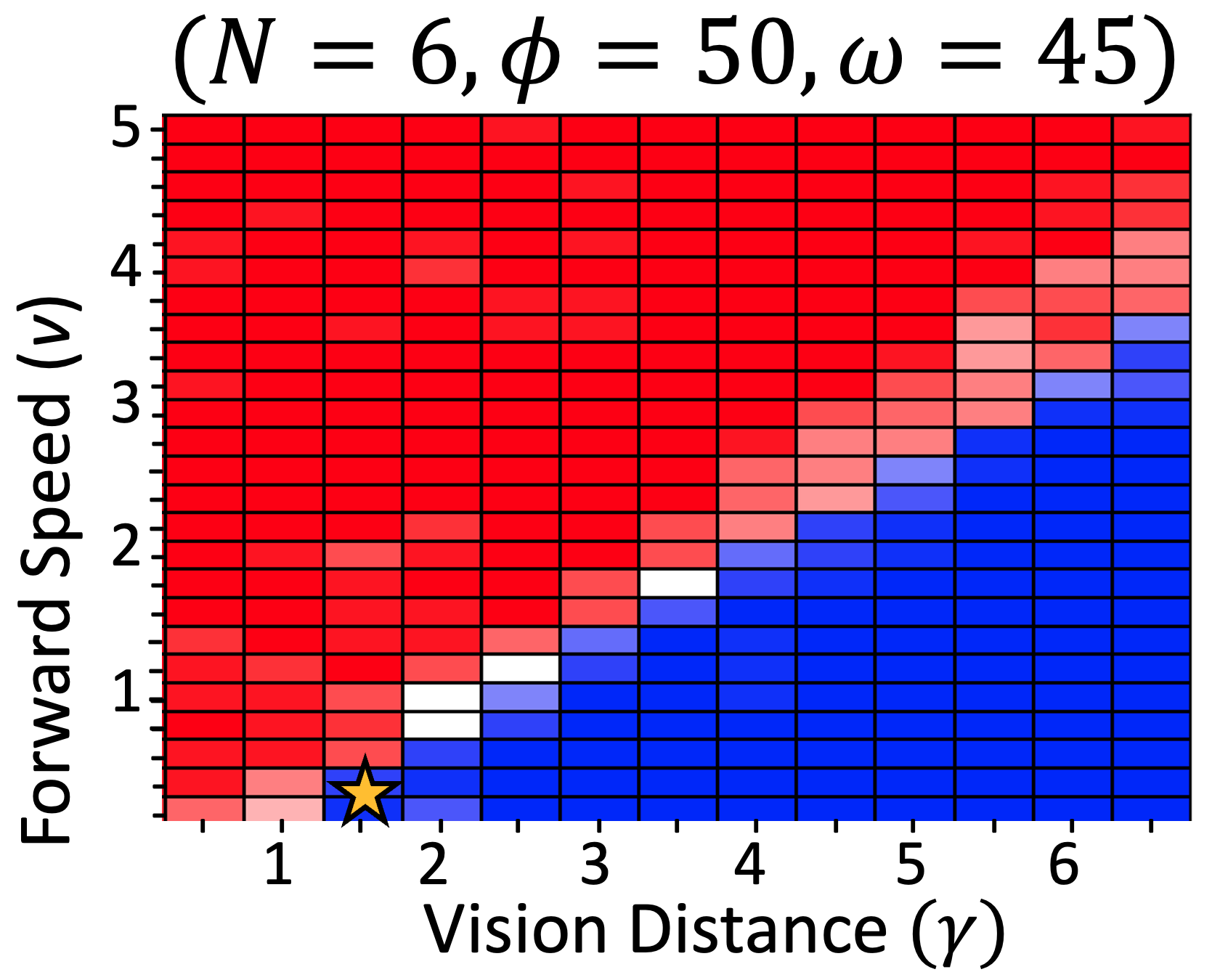}}
    \caption{Multiple phase diagrams generated by simulating agents with controller~\eqref{eq:mill_control} under various parameters.
    Blue regions indicate where milling occurs, while red regions correspond where milling was not achieved. Parameters combinations tested on real robots are marked with gold stars.
    }
    \Description{Multiple phase diagrams generated by simulating agents with milling controller under various parameters.}
\label{fig:multiple_phase_diagrams_mill}
\end{figure*}

Multiple phase diagrams have been generated and are shown in Figure~\ref{fig:multiple_phase_diagrams_mill}. As described in Section~\ref{sse:phase_approach}, the coloring of each diagram indicates the frequency with which the milling behavior occurred across the ten simulation trials. While we cannot yet guarantee that these phase diagrams are the most informative or reveal all parameter relationships, they nevertheless highlight key dependencies. For instance, holding all other parameters constant, the phase diagram of $\phi$ versus $N$ (Figure~\ref{fig:multiple_phase_diagrams_mill}) exhibits a narrow band in which milling occurs, suggesting a strong relationship between $N$ and $\phi$.

\subsubsection{Diffusion}
From our observations, a system of agents with the controller~\eqref{eq:diff_controller} does not produce as many different behaviors as a system of agents using the controller~\eqref{eq:mill_control}. Besides moving in an unorganized motion, the only observable behaviors that occur are diffusion and semi-diffusion, where agents aren't evenly spaced out.

\begin{figure*}[ht]
    \centering
    \subfigure[]{\includegraphics[width=.33\linewidth]{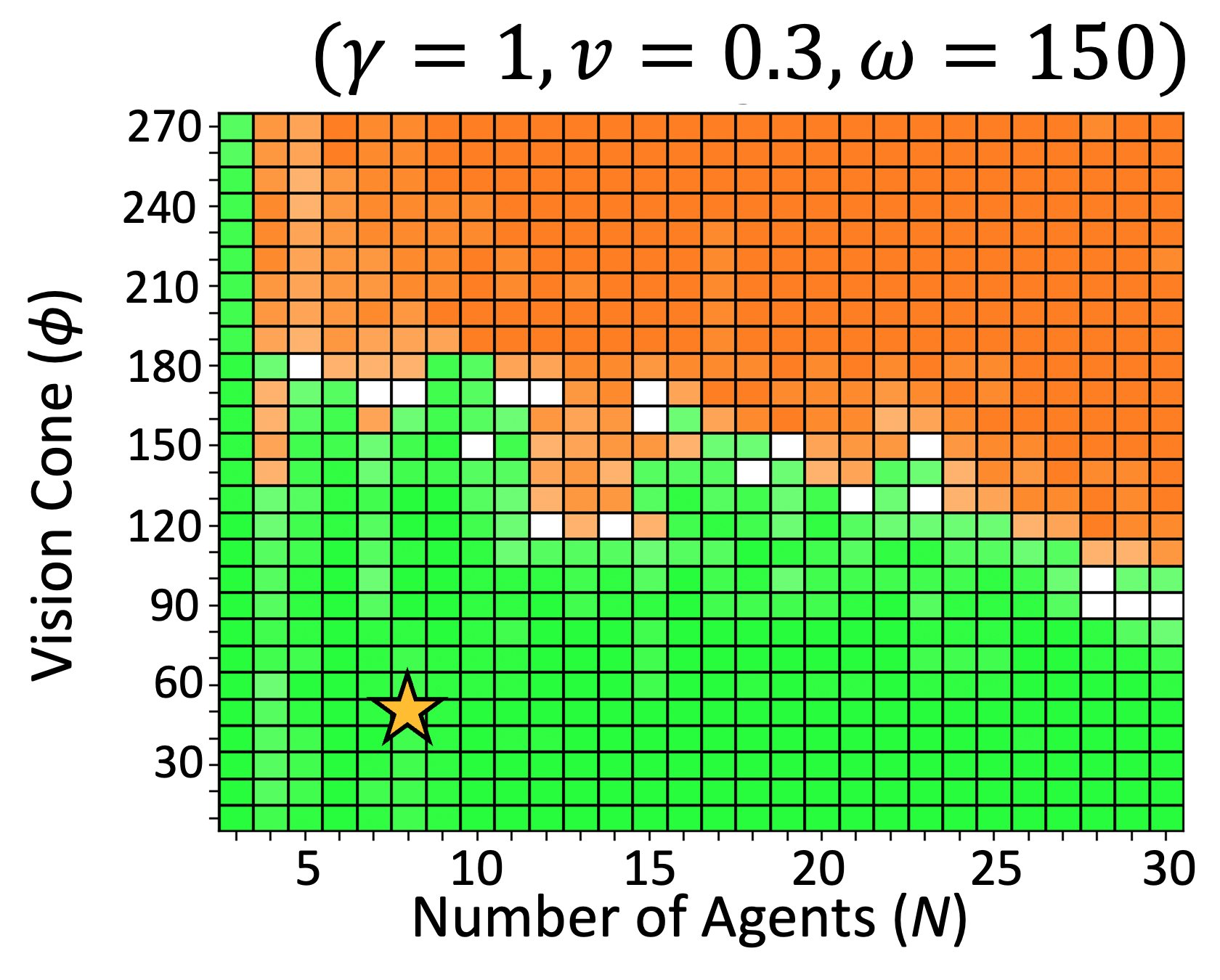}}
  \subfigure[]{\includegraphics[width=.33\linewidth]{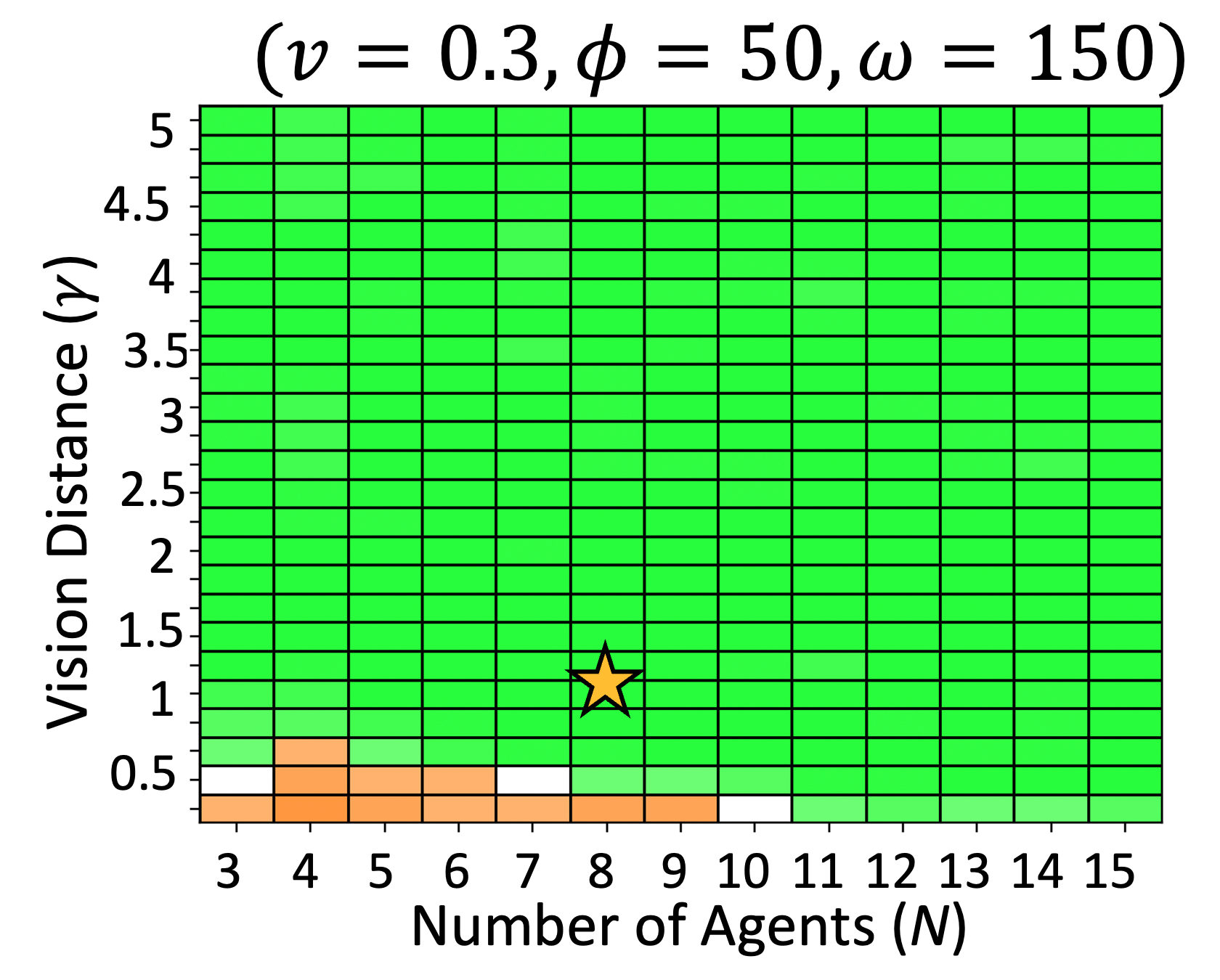}}
  \subfigure[]{\includegraphics[width=.33\linewidth]{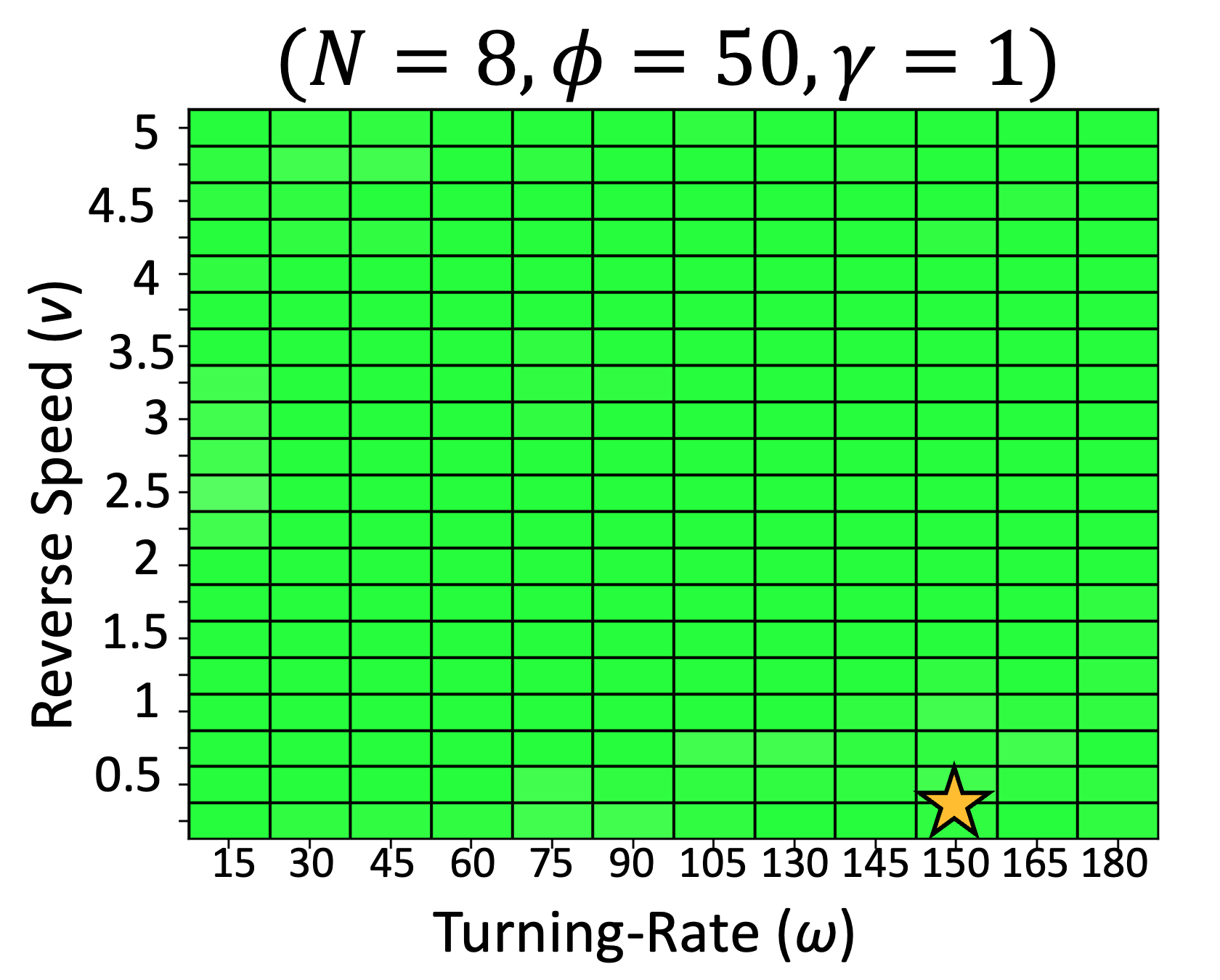}}
    \caption{Multiple phase diagrams generated by simulating agents with the ~\eqref{eq:diff_controller} controller. The green areas represent where diffusion occurred according to our constraints $\eta^2$ and the orange represent where it did not. Parameters combinations tested on real robots are labeled with stars.
    }
    \Description{Multiple phase diagrams of the diffusion behavior}
    \label{fig:multiple_phase_diagrams_diffusion}
\end{figure*}

The phase diagrams in Figure~\ref{fig:multiple_phase_diagrams_diffusion} are again only 2D slices of a more complex space; however, they show the regions where diffusion reliably occurs. There appears to again be a strong relationship between $N$ and $\phi$ whereas diffusion occurs no matter the speed $v$ and turning rate $\omega$.

\section{Robot Validation}\label{se:validation}
Our ultimate goal with this novel framework is to combine simulations with insights from swarm chemistry to deploy real robot swarms exhibiting predictable emergent behaviors. Naturally, there is a substantial gap between theoretical or simulated models and real-world robots. Real robots operate as discrete systems with finite sensing and actuation rates, meaning that time sampling (ignored in our continuous-time model) may affect behavior. In addition, sensing is often imperfect, producing false positives or negatives, and individual robots may differ slightly from one another. Even in a swarm of ``identical" robots, these idiosyncrasies prevent perfect reliability in behavior. Nevertheless, by following the methods outlined in~\cite{RV-KZ-CM-DSB-CN:24}, this gap can be sufficiently reduced to allow the behaviors observed in simulation to manifest on real robots. 

Experiments were deployed using TurboPis, which are 19x16x14cm robot controlled via Raspberry Pi 4 and have four Mecanum wheels that give them omni-directional locomotive capabilities as well as two servo motors that allow the onboard camera to pan and tilt giving it a controllable larger Field-of-View (FOV). Additional sensors include an ultrasonic range-finder and a 4 channel line tracker. Although fitted with various sensors and more complex computation capabilities, our aim is to study the potential of non-symbolic controllers, so for our purposes we only use one sensor (RGB camera) that generates a binary output of 1 when anything green is detected or 0 otherwise. We also kept the agent dynamics simple and had the robots operate under a unicycle model where the robot could only move forward or backward and/or rotate its heading similar to~\eqref{eq:simple_kinematics}.

Naturally, it is easier to tune various parameters in simulation than with real hardware. For instance, it is not easy to simply tune the sensing capabilities of the sensors built into the TurboPis and are limited to only being able to use up to~$N=16$ of the robots we had available. Nevertheless, the phase diagrams shown in Figure~\ref{fig:multiple_phase_diagrams_mill} and Figure~\ref{fig:multiple_phase_diagrams_diffusion} can serve as guides for deploying the TurboPis to reproduce the corresponding behaviors observed in simulation.

After applying the Real-to-Sim-to-Real (RSRS) process~\cite{RV-KZ-CM-DSB-CN:24}, we determined that the TurboPis have a maximum forward and reverse speed of $0.3~\mathrm{m/s}$ and a maximum turning rate of $150~\mathrm{deg/s}$. Each TurboPi can detect other robots by sensing green balls placed on top of them. The robots operate at a sampling frequency of $40 Hz$, with a reliable maximum vision distance of $1.1 m$ and a field of view of $50\degree$.

Given these measured capabilities, we can find points from the phase diagrams in Figure~\ref{fig:multiple_phase_diagrams_mill} and Figure~\ref{fig:multiple_phase_diagrams_diffusion} within the regions where we see the desired behaviors occur. For milling, we chose to run $N = 6$ robots with a set forward speed of $v=0.25 \frac{m}{s}$, and turning rate of $\omega = 45 \frac{deg}{s}$ (this point is illustrated with a star in Figure~\ref{fig:multiple_phase_diagrams_mill}). In our attempt to produce the diffusing behavior, we chose to run $N=8$ robots, reverse speed of $v = 0.3 \frac{m}{s}$ and turning rate of $\omega = 150\frac{deg}{s}$ (this point is also illustrated with a star in Figure~\ref{fig:multiple_phase_diagrams_diffusion}).

\begin{figure}[h]
    \centering
    \subfigure[]{\includegraphics[width=.96\linewidth]{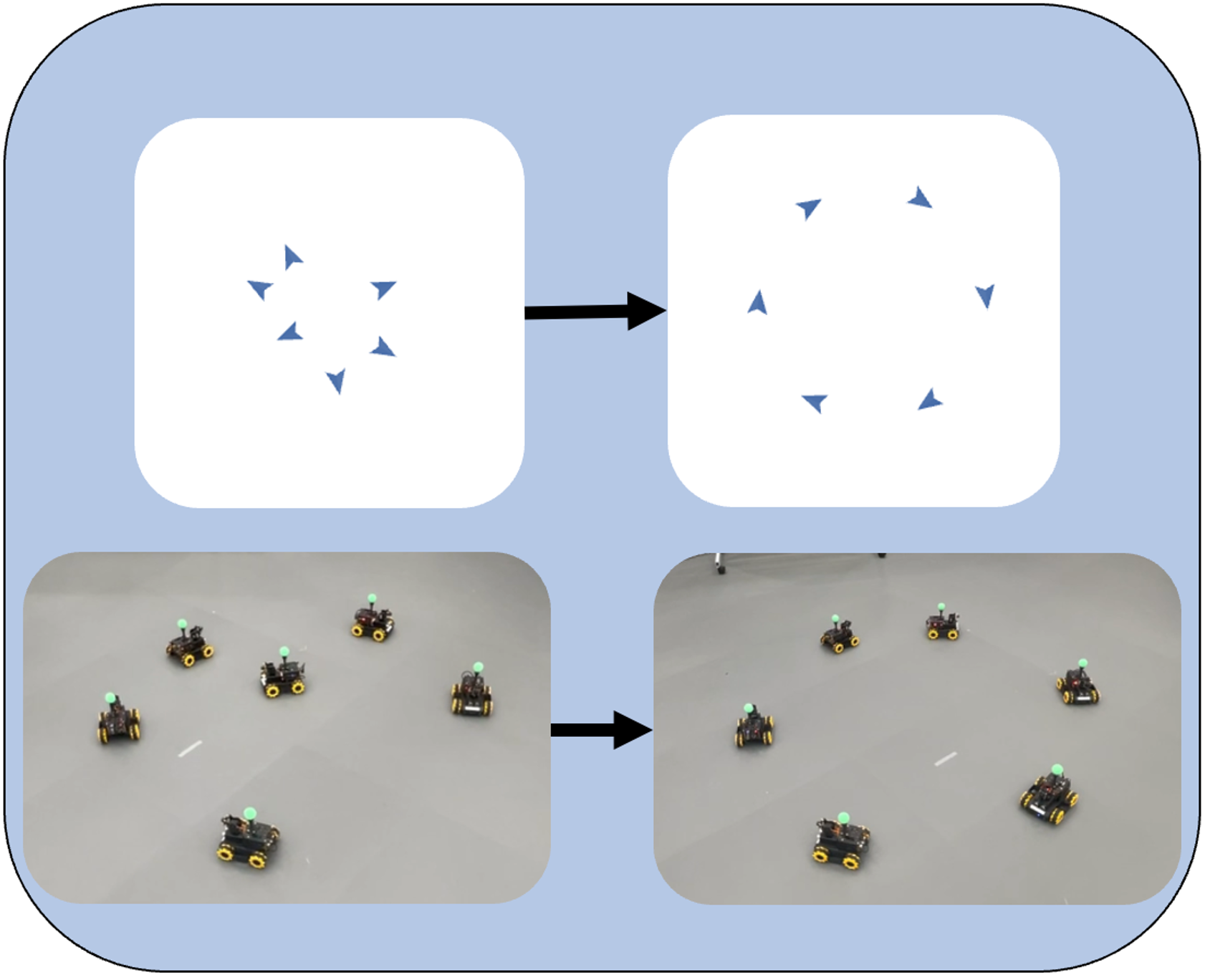}}
  \subfigure[]{\includegraphics[width=.96\linewidth]{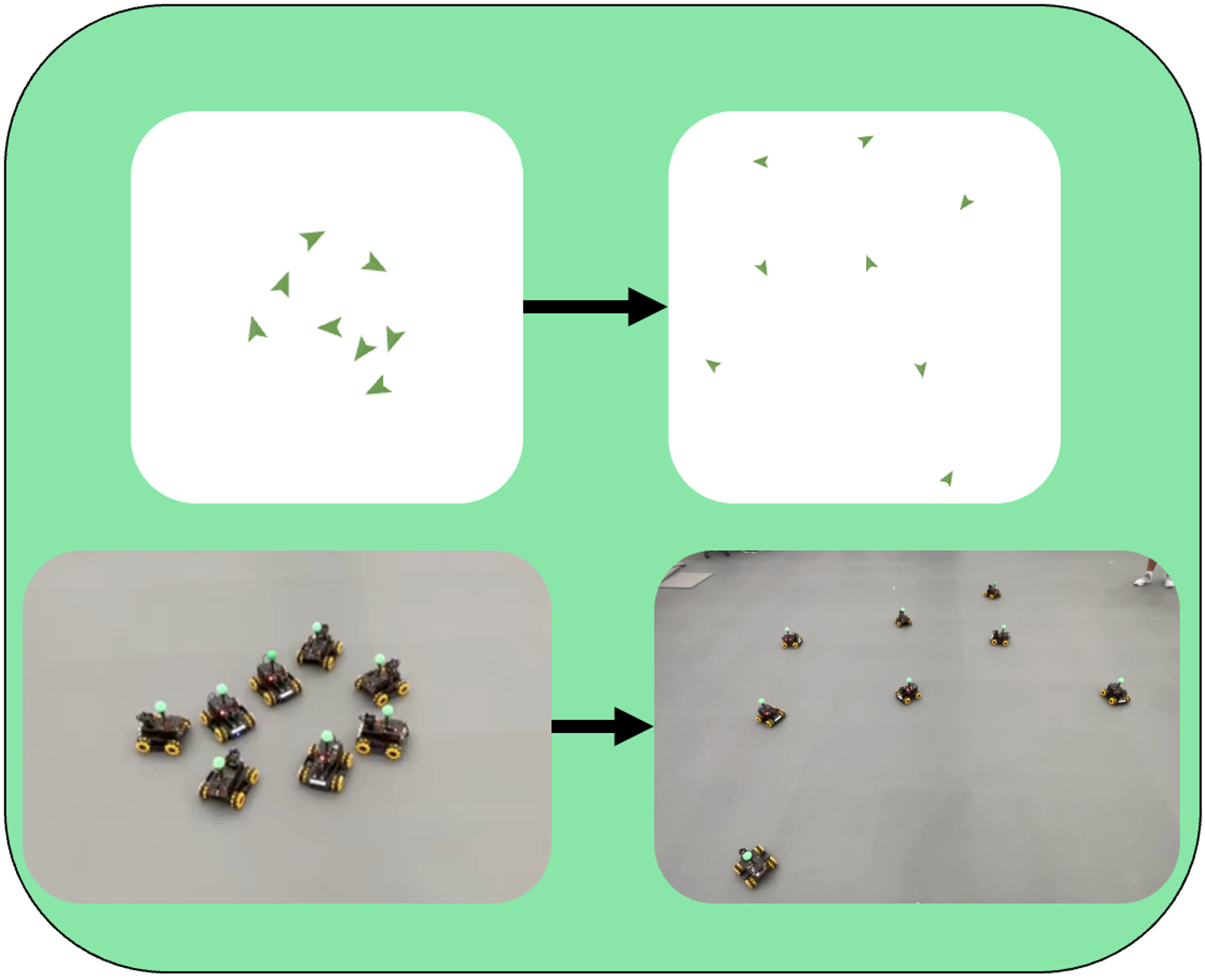}}
    \caption{(a) Start and finish of six agents milling in NetLogo simulator and six TurboPi successfully milling using the controller~\eqref{eq:mill_control} with $v = 0.25\frac{m}{s}$, $\omega = 45 \frac{deg}{s}$, $\gamma = 1 m$, and $\phi = 50 \degree$. (b) Start and finish of eight agents diffusing in NetLogo simulator eight TurboPi successfully diffusing using the controller~\eqref{eq:diff_controller} with $v = 0.3\frac{m}{s}$, $\omega = 150 \frac{deg}{s}$, $\gamma = 1 m$, and $\phi = 50 \degree$.}
    \label{fig:sim-real-experiments}
    \Description{Comparison of milling and diffusion behaviors being produced in simulation and on real robots }
\end{figure}

As shown in Figure~\ref{fig:sim-real-experiments}, the experiments successfully reproduced the intended behaviors under the selected parameter settings. Additional trials conducted within the regions identified as successful in the phase diagrams also achieved a high rate of behavioral replication on the real robots. However, the milling behavior exhibited a narrower range of stability in real-world conditions. While simulations predicted stable milling with $N=7$ robots (with all other parameters unchanged), the TurboPis initially formed a milling circle that gradually broke apart as the experiment progressed. This discrepancy suggests that further refinement of the simulator is needed to better capture the imperfections present in the physical robots. In contrast, the diffusion behavior proved far more robust, consistently emerging across all experimental trials.

\section{Conclusions}\label{se:conclusions}
This work introduced a novel framework for analyzing and modeling swarm systems that bridges the exploratory strengths of agent-based modeling with the analytical rigor of traditional engineering. By integrating these complementary perspectives, the framework offers a structured means to understand how minimal local rules and simple sensing mechanisms can give rise to complex emergent behaviors given the right parameters. In doing so, it provides a foundation for the systematic design, prediction, and eventual deployment of efficient and scalable robot swarms.

Future work will focus on extending this framework to include heterogeneous swarms composed of multiple agent “species,” as in the original Swarm Chemistry~\cite{HS:09}, to explore new classes of emergent behaviors. The same analysis can then be applied to identify regions of parameter space where these behaviors occur reliably. In addition, since the current phase diagrams were manually generated, based on systematic exploration of parameter variations, an important next step will be to develop automated techniques for exploring parameter spaces. Such methods could more efficiently identify regions of consistent behavior and determine which phase diagrams yield the most informative insights.






\bibliographystyle{ACM-Reference-Format} 
\bibliography{ricardo}


\end{document}